\documentclass{article}

\usepackage[final]{neurips_2019}

\usepackage[utf8]{inputenc} 
\usepackage[T1]{fontenc}    
\usepackage{hyperref}       
\usepackage{url}            
\usepackage{booktabs}       
\usepackage{amsfonts}       
\usepackage{nicefrac}       
\usepackage{microtype}      
\usepackage{graphicx}
\usepackage{enumitem}
\setlist{nolistsep,leftmargin=*}

\newcommand{\doi}[1]{\href{https://dx.doi.org/#1}{doi:#1}}
\renewcommand{\subsection}[1]{\textbf{#1}\enspace}

\title{Deep learning predictions of sand dune migration}

\author{%
  Kelly Kochanski$^{1,2}$, Divya Mohan$^{2,3}$, Jenna Horrall$^{2,4}$, Barry Rountree$^2$ and Ghaleb Abdulla$^2$ \\ \footnotesize{$^1$University of Colorado Boulder $^2$Lawrence Livermore National Laboratory}\\ \footnotesize{$^3$University of California Berkeley $^4$James Madison University}
}

\begin{document}

\maketitle

\vspace{-30pt}

\section{The problem with dunes}
A dry decade in the Navajo Nation has killed vegetation, dessicated soils, and released once-stable sand into the wind.
This sand now covers one-third of the Nation's land, forming dunes that move 40~m per year (Fig. 1a) across roads, gardens and grazing land, threatening hundreds of homes~[1].
Many arid regions have similar problems: sand dunes encroach on solar panel installations in the Mojave Desert,
and global warming has increased dune movement across farmland in Namibia and Angola [2].
As global warming continues, these areas will likely get warmer, drier [3], and sandier.

\subsection{Limitations of current models}
With enough warning, people can protect their property from sand dunes.
Current dune models, unfortunately, do not scale well enough to provide useful forecasts for the $\sim$5\% of land surfaces covered by mobile sand.
Dunes are shaped by nonlinear processes, such as turbulence and granular motion [4,5], that are expensive to compute.
For example, modelling the motion of sand dunes across the entire Navajo Nation (70,000 km$^2$) with leading dune model ReSCAL [6] would take $\sim$200 years.
Dune predictions also scale poorly in a labor sense, as existing models require knowledge of hard-to-measure variables like sand grain size (see efforts in~[7]).

\subsection{Why deep learning could do better}
In the past few years, deep learning has been shown to perform well in the prediction of spatio-temporal patterns, such as videos [8,9,10]. 
Dune fields are a classic example of a self-organised spatio-temporal pattern [11], and we hypothesise that they would be modelled well by these techniques, at a low computational cost compared to current models.

\subsection{Aim}
We propose to test the ability of two deep learning algorithms to model the motion of dunes.
Our target is to make regional forecasts that give individuals 1 year of warning before their buildings, farms, or homes are covered by dunes.
We estimate that this requires a model 10$^5\times$ faster than existing physics-based models, which predicts dune-covered areas with $>$80\% accuracy.

\section{Building a scalable dune model with deep learning}

\begin{figure}
  \centering
  \includegraphics[width=\textwidth]{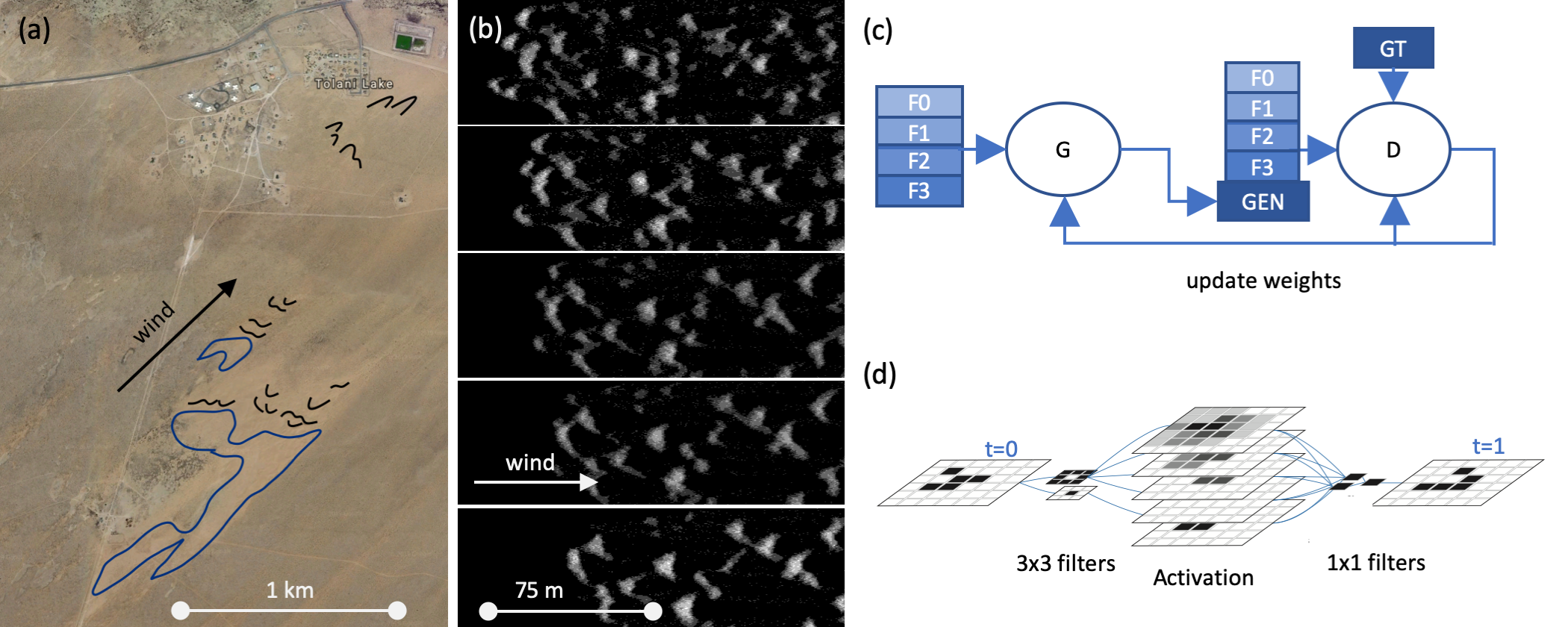}
  \caption{(a) Movement of sand dunes towards Tolani Lake, AZ, USA;
  the dune fields from 2000 are circled in blue, with dunes that now exceed that area marked in black; 
  (b) simulated sand dune data, showing elevations of simulated dunes over a 9 month period; 
  (c) GAN training workflow, showing input of 4 simulated frames F0--F3 into generator G and discriminator D; 
  (d) architecture of a CNN designed to represent cellular automata.}
\end{figure}

We will train the model on simulated data, then test it against observations.
Training on a state-of-the-art simulation will give us access to clean and abundant data, and validate our work against a standard accepted by the geoscience community.

\subsection{Data generation} We created a simulated dataset with 360 dune time series. Each time series consists of elevation maps
covering 75$\times$300 m areas with 0.5 m$^2$ pixels every 3.0 days for 0.5 years (see Fig. 1c--d).
The simulations were run with physics-based dune simulation Rescal-snow [12], a streamlined adaptation of an older simulation (ReSCAL, [6]) used in ground-breaking dune studies~[13,14].
This data structure maximises the data's resemblance to both black-and-white video data and to satellite-generated height maps of real dunes.

Our eventual goal is to build a model that works on dunes in many locations.
This means that our final model must be trained on data that covers a wide range of wind patterns, dune geometries, and sand grain shapes, sizes, and densities,
and must learn the effects of these physical parameters using height map data alone.
For initial training, we simplified the problem by generating dunes built only of fine-grained quartz sand blown by a constant 10 m/s wind, as in [15]. We will increase the complexity of the training data gradually, by simulating a wider range of conditions. 
This process will echo the work of the geoscience community, who have increased the complexity of their models by one variable at a time.

\subsection{Deep learning model design}
We selected two deep learning models with contrasting approaches:
\begin{enumerate}
\vspace{-5pt}
\item  A video prediction GAN, based on a multi-scale CNN which uses alternating convolutions and rectified linear units to generate 32$\times$32 pixel frames based on four frames of history (Fig. 1c). For full details see [10].

\item A CNN designed to emulate the algorithmic structure of the physics-based dune simulation. Specifically, our simulation is a cellular automaton, a structure based on discrete nearest-neighbor interactions. [16] showed that some cellular automata may be represented by a CNN with 3$\times$3 convolutions (Fig. 1d). See full details in [16] and discussion in \textsection\ref{sec: appendix}.
\end{enumerate}

\subsection{Preliminary work} 
Once trained, these models generate frames respectively $10^7$ and $10^4$ times faster than the physics-based simulation. Read more in \textsection\ref{sec: appendix}.

\subsection{Learning from observational data}
Height maps of real dune fields (called `digital elevation models') are widely available: see the list of sources in [17]. We will test our model against historical data from these fields.
If the model does not performs well on simulated data but not on observations,
we will continue to improve the model by training on observational data.

\subsection{Final product} We will apply our model to satellite-generated height maps of dune fields to create forecasts for vulnerable regions including the Navajo Nation and the Kalahari Desert.

\section{Conclusion}
We propose to build a deep learning model of sand dunes. If successful, the model will enable us to make the first large-scale quantitative predictions of dune migration,
and to give actionable information to people living through desertification.
Our team contains both computer scientists and geologists, so we have the skills to scale our model and to connect it with existing scientific work.

This work will also pilot the use of deep learning for modelling spatio-temporal patterns in nature.
Sand dunes are spatio-temporal patterns visible in image-like satellite data. They affect a large fraction of the Earth, and many people would benefit from scalable predictions. Current models, however, are sorely limited by their computational expense, do not scale, and have not yet taken great advantage of satellite data.
These features are shared between sand dunes and other climate problems, including river incision [18], snow drifts~[19], storm tracks [20] and deltas subsiding to the sea.

Deep learning could be a valuable new tool for accelerating models of these natural hazards,
and for integrating new data into those models. The scope of scientific spatio-temporal problems which can be modeled with deep learning, however, is not yet defined. Broadening that scope it could enable us to make high-resolution natural hazard predictions at global scale.

\subsubsection*{Acknowledgments}

This research was supported by a Department of Energy Computational Science Graduate Fellowship (DE-FG0297ER25308). This work was performed under the auspices of the US Department of Energy by Lawrence Livermore National Laboratory under Contract DE-AC52007NA27344.

\vspace{-10pt}

\section*{References}

\small

[1] Reedster, M.H. \& Wessells, S.M. \ (2017) A record of change --- science and elder observations on the Navajo Nation: U.S. Geological Survey General Information Product 181, video, \doi{10.3133/gip181}

[2] Thomas, D. S., Knight, M., \& Wiggs, G. F. (2005). Remobilization of southern African desert dune systems by twenty-first century global warming. Nature, 435(7046), 1218. \doi{10.1038/nature03717}

[3]  IPCC (2014). Climate Change 2014: Synthesis Report. Contribution of Working Groups I, II and III to the Fifth Assessment Report of the Intergovernmental Panel on Climate Change [Core Writing Team, R.K. Pachauri and L.A. Meyer (eds.)]

[4] Bagnold, R. (1941) "The Physics of Blown Sand and Desert Dunes." Springer Netherlands, ISBN 978-94-009-5682-7

[5] Kok, J. F., Parteli, E. J., Michaels, T. I., \& Karam, D. B. (2012). The physics of wind-blown sand and dust. Reports on progress in Physics, 75(10), 106901.

[6] Rozier, O., \& Narteau, C. (2014) A real‐space cellular automaton laboratory. Earth Surface Processes and Landforms 39(1) 98--109, \doi{10.1002/esp.3479}

[7] Bogle, R., Redsteer, M. H., \& Vogel, J. (2015). Field measurement and analysis of climatic factors affecting dune mobility near Grand Falls on the Navajo Nation, southwestern United States. Geomorphology, 228, 41-51. \doi{10.1016/j.geomorph.2014.08.023}

[8] Lee, A. X., \ Zhang, R. \ et al. (2018) Stochastic adversarial video prediction. ArXiv preprint arXiv:1804.01523

[9] Simonyan, K., \ \& Zisserman, A. (2014) Two-stream convolutional networks for action recognition in videos. Advances in neural information processing systems arXiv:1406.2199

[10] Mathieu, M., \ Couprie, C. \ \& LeCun, Y. (2016) Deep multi-scale video prediction beyond mean-squared error. ICLR, arxiv:1511.05440

[11] Ewing, R. C. \ Kocurek, G. \ \& Lake, L. W. (2006) Pattern analysis of dune-field parameters. Earth Surface Processes and Landforms, 31, 1176--1191. \doi{10.1002/esp.1312}

[12] Kochanski, K., \ Defazio, G. \ Green, E. \ et al. (2019) Rescal-snow: A model of dunes and snow waves. \href{https://github.com/kellykochanski/rescal-snow}{https://github.com/kellykochanski/rescal-snow}

[13] Zhang, D., Narteau, C., Rozier, O., \& Du Pont, S. C. (2012). Morphology and dynamics of star dunes from numerical modelling. Nature Geoscience, 5(7), 463.

[14] Gao, X., Narteau, C., Rozier, O., \& Du Pont, S. C. (2015). Phase diagrams of dune shape and orientation depending on sand availability. Scientific reports, 5, 14677.

[15] Narteau, C., Zhang, D., Rozier, O., \& Claudin, P. (2009). Setting the length and time scales of a cellular automaton dune model from the analysis of superimposed bed forms. Journal of Geophysical Research: Earth Surface, 114(F3).

[16] Gilpin, W. (2018). Cellular automata as convolutional neural networks. ArXiv preprint, arXiv:1809.02942

[17] Hugenholtz, C. H., Levin, N., Barchyn, T. E., \& Baddock, M. C. (2012). Remote sensing and spatial analysis of aeolian sand dunes: A review and outlook. Earth-science reviews, 111(3-4), 319-334. \doi{10.1016/j.earscirev.2011.11.006}

[18] Tucker, G. E., \& Hancock, G. R. (2010). Modelling landscape evolution. Earth Surface Processes and Landforms, 35(1), 28-50. \doi{10.1002/esp.1952}

[19] Kochanski, K., Anderson, R. S., \& Tucker, G. E. (2019). The evolution of snow bedforms in the Colorado Front Range and the processes that shape them. The Cryosphere, 13(4), 1267-1281 \doi{https://doi.org/10.5194/tc-13-1267-2019}.

[20] Kim, S., Kim, H., Lee, J., Yoon, S., Kahou, S. E., Kashinath, K., \& Prabhat, M. (2019, January). Deep-hurricane-tracker: Tracking and forecasting extreme climate events. In 2019 IEEE Winter Conference on Applications of Computer Vision (WACV) (pp. 1761-1769). IEEE.

\clearpage

\appendix
\section{Preliminary results}
\label{sec: appendix}

\subsection{GAN results}
The preliminary results from the GAN (Figs.~2a--b) are promising, though the model is not yet fully trained and tuned and has not yet reached a usable level of accuracy.
As shown in Fig.~1c, the model takes in a series of four images (F0--F3) and generates a fifth (GEN).
These frames may be separated by any number of simulation time steps; we tested intervals of 3--90 days and found that performance was best for intervals of 35--71 days. We believe this is the time frame in which all dunes move at least one pixel,
but no dunes move across the entirety of the 32$\times$32 tiles used by the convolutions in the generator.
This leads to a model that is quite computationally efficient.
Using the 68-day time steps shown in Figs.~2a--b, the model generates frames 10$^7\times$ faster than the physics-based simulation.

We expect the accuracy of the model to improve with additional training and tuning.
At present, the model captures the rough shapes of dunes and their downwind motion (Fig.~2a), but it performs less well on fields of small dunes,
which frequently merge, branch, and change shape (Fig.~2b).

\begin{figure}
  \centering
  \includegraphics[width=\textwidth]{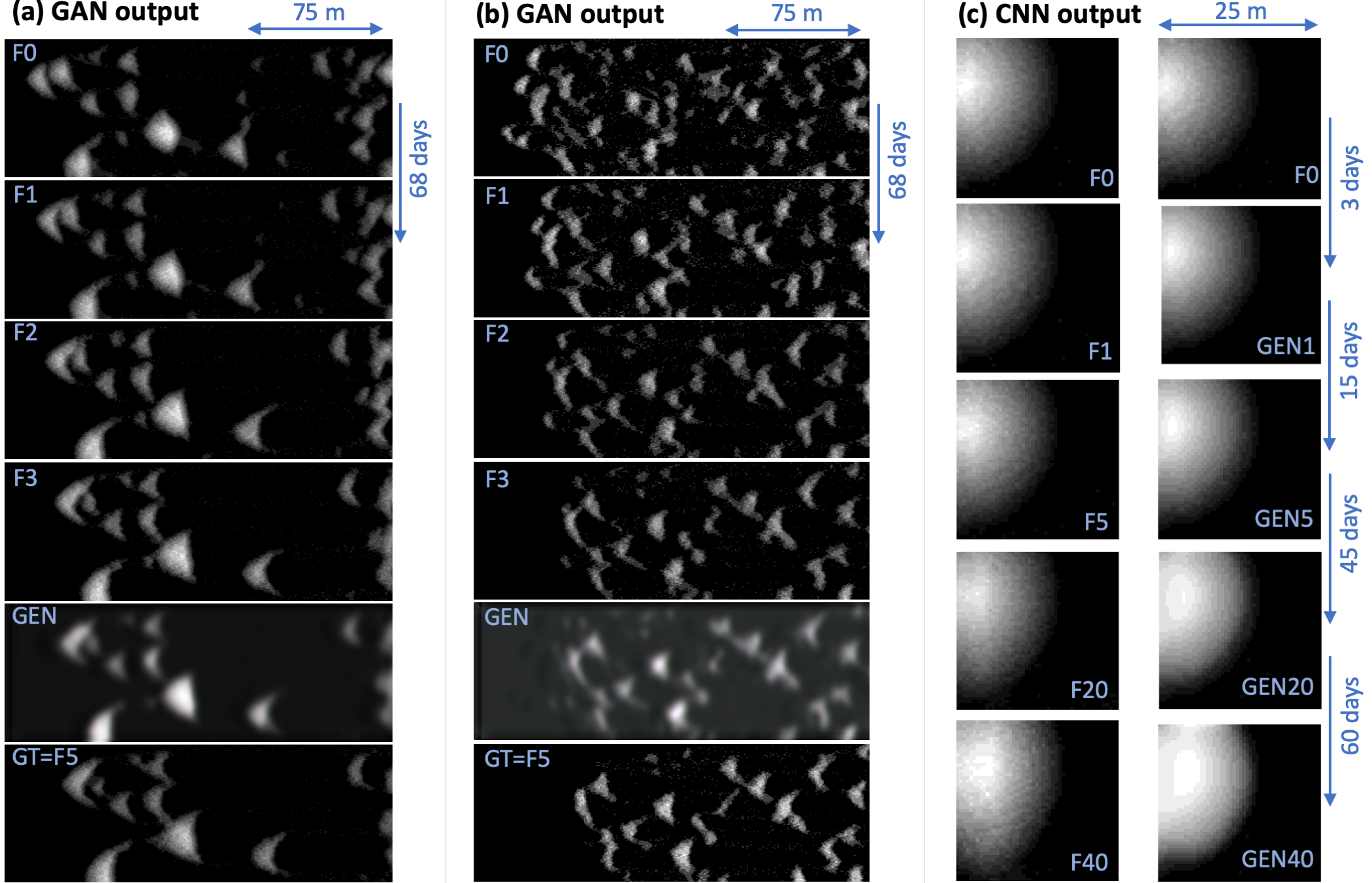}
  \caption{(a, b) Example frames generated by the GAN; (c) comparison between time series generated by the physics-based simulation (left) and CNN (right). F0--F40 are images generated by the physics-based simulation. GEN frames are generated. GT are ground truth.}
\end{figure}

\subsection{Comparison of CNN and physics-based simulation architectures}
The CNN we tested (Fig.~1d) was designed to learn the rules of a cellular automaton (CA) as in [16].
The physics-based dune model we used is a cellular automaton [6], so
this gave us an opportunity to work with a relatively simple and elegant neural network,
and to test whether the result in [16] is applicable to a scientific CA.

This CNN in [16], however, was tested on a much simpler CA than a sand dune: Conway's Game of Life. 
'Life' contains binary (black/white) cells on a 2D grid. 
The cells evolve at regular intervals according to the natures of their nearest neighbours. [16] shows that 'Life' may thus be represented perfectly by a CNN with nearest-neighbour (3$\times$3) convolutions.
The dune CA, in contrast, contains 30 cell states, which represent various grain movements and wind flow directions, on a 3D grid [6]. The cells evolve stochastically, usually according to the natures of their nearest neighbours, but occasionally due to distant effects (avalanches).
Thus, our use of the CNN tests the extent to which its simplified structure
is an acceptable approximation of the dune CA.

To complete this test thoroughly, we built variations of the CNN in [16] which re-introduced some of the complexity of the dune CA: (1) a 2D map with continuous states representing different dune heights, (2) a 3D map with binary grain/fluid states, (3) a 3D map with continuous states representing various wind directions, and (4--6) all of the above with added 1-pixel noise. 
We did not add long-range convolutions as we felt that those were effectively covered by the multi-scale CNN in the GAN. Below, we show results from the 2D case (configuration 1), as it resembles satellite data most closely and would most useful for real dunes.

\subsection{CNN results}
Our preliminary results with the CNN (Fig.~2c) indicate that it is not a useful representation of the sand dune simulation.
First, it is not as fast as we'd hoped: it generates frames 10$^4\times$ faster than the physics-based simulation. This might be improvable with better hardware or optimised data processing.

Second, its accuracy is poor, and the errors it makes are consistent with the dissimilarities discussed above between the CNN architecture and the physics-based simulation discussed above.
For example, the CNN tends to sort dunes into flattened layers (Fig.~2b).
It also imposes symmetry on the dunes: Fig.~2c shows that, when the simulation is applied repeatedly, the CNN increases the symmetry of a small sand hill, where the physics-based simulation stretches it slightly into an elongate dune.
We suspect that this symmetry occurs because the CNN over-simplifies the cell states, and removes the details of the flow field that induce movement in real dunes.
As the time step of the CNN is limited to the 3-day time step of the physics based simulation, our target year-long prediction would require the generation of 122 future frames. We expect that the cumulative error over this period would grow very large.

\end{document}